\documentclass[10pt,twocolumn,letterpaper]{article}

\usepackage{cvpr}
\usepackage{times}
\usepackage{epsfig}
\usepackage{graphicx}
\usepackage{amsmath}
\usepackage{amssymb}
\usepackage{flushend}
\usepackage{subcaption}
\usepackage{multirow}
\usepackage{bm}

\usepackage[bottom,flushmargin,hang,multiple]{footmisc}
\usepackage{booktabs}
\usepackage{comment}
\usepackage{authblk}

\usepackage{amsmath,amsthm}
\usepackage{amssymb, color}
\usepackage{float}
\graphicspath{{figure/}}

\renewcommand{\phi}{\varphi}

\usepackage{mathtools}

\usepackage[pagebackref=false,breaklinks=true,letterpaper=true,colorlinks,bookmarks=false]{hyperref}

\cvprfinalcopy % *** Uncomment this line for the final submission

\def\cvprPaperID{6172} % *** Enter the CVPR Paper ID here

\ifcvprfinal\fi
\begin{document}

%%%%%%%%% TITLE
\title{Image Scene Graph Generation (SGG) Benchmark}

% \author{Xiaotian Han \and Jianwe Yang \and Houdong Hu \and Lei Zhang \and Jianfeng Gao \and Pengchuan Zhang }
\author[1]{Xiaotian Han}
\author[2]{Jianwei Yang}
\author[1]{Houdong Hu}
\author[3]{Lei Zhang}
\author[2]{Jianfeng Gao}
\author[2]{Pengchuan Zhang}
\affil[1]{Microsoft Cloud + AI}
\affil[2]{Microsoft Research at Redmond}
\affil[ ]{\textit \{xiaothan, jianwyan, houhu, jfgao, penzhan\}@microsoft.com, \space leizhang.china@gmail.com}

\renewcommand\Authands{ and }

\maketitle

% \blfootnote{$*$ indicates equal contributions.}
\begin{abstract}
  There is a surge of interest in image scene graph generation (object, attribute and relationship detection) due to the need of building fine-grained image understanding models that go beyond object detection. 
  Due to the lack of a good benchmark, the reported results of different scene graph generation models are not directly comparable, impeding the research progress. %, even confusing. 
  We have developed a much-needed scene graph generation benchmark based on the maskrcnn-benchmark\cite{massa2018mrcnn} and several popular models.
  This paper presents main features of our benchmark and a comprehensive ablation study of scene graph generation models using the Visual Genome\cite{krishnavisualgenome} and OpenImages Visual relationship detection\cite{OpenImages2} datasets. Our codebase is made publicly available at \url{https://github.com/microsoft/scene_graph_benchmark}.
  % In addition, 
  % by adding an attribute detection head to the scene graph generation model, 
  % we propose a universal image understanding model that achieves object-attribute-relationship detection in a single model. 
\end{abstract}

\vspace{-5mm}
\section{Introduction}
\label{sec:intro}
Understanding a visual scene goes beyond recognizing individual objects in isolation. Visual attributes of an object and relationships between objects also constitute rich semantic information about the scene. Scene Graph Generation (SGG)~\cite{Xu_2017} is a task to detect objects, their attributes and relationships using scene graphs~\cite{krishnavisualgenome}, a visually-grounded graphical structure of an image. SGG has proven useful for building more visually-grounded and compositional models for other computer vision and vision-language tasks, like image/text retrieval~\cite{Johnson_2015_CVPR,wang2020cross}, image captioning~\cite{gao2018image,yang2019auto,gu2019unpaired}, visual question answering~\cite{ben2019block,ghosh2019generating}, action recognition~\cite{ji2019action}, image generation~\cite{johnson2018image,sylvain2020objectcentric}, and 3D scene understanding~\cite{armeni20193d}. 

\begin{table*}[t]
\begin{center}
%\scriptsize
\resizebox{\linewidth}{!}{
\begin{tabular}{ c|c|c|c|c|c}
\toprule
 Code base & Supported datasets & Supported methods & \shortstack{Last commit time \\ (Pytorch version)} & \shortstack{Attribute \\ detection} & \shortstack{OD/SGG \\ decouple-able} \\
\midrule
Neural Motif \cite{Zellers_2018} & VG & IMP,NM & 11/07/2018 (0.3) & No & No  \\
RelDN \cite{zhang2019graphical} & VG, OI & NM,GRCNN,RelDN & 08/19/2019 (1.0) & No & Yes  \\
GRCNN \cite{Yang_2018} & VG & IMP,MSDN,GRCNN & 03/31/2020 (1.0) & No & No  \\
SGG-Pytorch \cite{tang2020sggcode} & VG & IMP,NM,VCTree~\cite{tang2019learning} & 12/13/2020 (1.2) & Yes & No  \\
SG Benchmark (ours) & VG, OI & IMP,MSDN,NM,GRCNN,RelDN & 06/07/2021 (1.7) & Yes & Yes  \\
\bottomrule
\end{tabular}
}
\end{center}
\vspace{-4mm}
\caption{Comparison of different open-sourced code bases for scene graph generation (SGG). Our code base supports two datasets (Visual Genome (VG) and Open Images (OI)) and five popular SGG methods. It is maintained regularly with up-to-date Pytorch support. It supports both attribute and relationship detection, and its object detection module (OD) and SGG module are decouple-able, which enables user to use customized OD models/results.}
\label{tab:sggrepos}
\vspace{-3mm}
\end{table*}

There are several open-sourced code bases and models for SGG, see, e.g., \cite{Zellers_2018,Yang_2018,zhang2019graphical,tang2020sggcode}. However, none of the existing code bases satisfies the following five criteria that from our perspective are important for easy use: (1) Support both Visual Genome (VG) and Open Images (OI) datasets, where VG is a popular academic testbed and OI is a large-scale public benchmark; (2) Support various popular methods, especially RelDN~\cite{zhang2019graphical} that we found is very effective in improving SGG performance; (3) Under continuous maintenance and up-to-date Pytorch support; (4) Support attribute detection, which is crucial in vision-language downstream tasks~\cite{anderson2018bottom,zhang2021vinvl}; and (5) Make the object detection (OD) module and the SGG module decouple-able, allowing users to use their customized detection models and results. Our SGG Benchmark code base, built on top of maskrcnn-benchmark~\cite{massa2018mrcnn}, satisfies all these five criteria and has several other useful features. We compare ours with existing SGG code bases in Table~\ref{tab:sggrepos} and list main features of our code base as below.
% In our SG benchmark repo, we also unify the evaluation metrics of SGG and make a comprehensive comparison of five most popular SGG methods. 

\begin{itemize}
    \item \textbf{Provide a generic SGG dataset class that supports both VG and OI datasets and is easily adaptable to customized SGG datasets.} We have integrated evaluation metrics from official Open Images relationship detection challenge for OpenImages Visual relationship detection\cite{OpenImages2}, and implemented widely used Visual Genome relationship detection Recall calculation for scene graph generation on the Visual Genome\cite{krishnavisualgenome}. 
    \item \textbf{Support multiple popular scene graph generation algorithms.} We have integrated five widely used scene graph generation algorithms into our framework, including iterated message passing (IMP) \cite{Xu_2017}, multi-level scene detection networks (MSDN) \cite{Li_2017}, graph R-CNN (GRCNN) \cite{Yang_2018}, neural motif (NM) \cite{Zellers_2018} and RelDN \cite{zhang2019graphical}.
    \item \textbf{Modular design with decouple-able object detection and SGG modules.} We have decomposed SGG framework into three components: object detection, attribute classification and relationship detection. One can easily replace or insert new components. Our framework decouples object detection and relationship detection so that one can do relationship detection with bounding boxes and labels from any object detector.
    \item \textbf{Fast and portable dataset format.} Our code base uses Tab Separated Values (TSV) data format, which is portable and flexible for data manipulations. We have also developed tools to visualize this data format directly. We will release the visualization tool later.
    \item \textbf{Large models with state of the art performance.} We have trained large models with ResNeXt-152 backbone for both object detection and scene graph generation. Our large models achieve state of the art performance on both Open Images relationship detection and Visual Genome scene graph generation.
    % \item We achieved state of the art results on both Open Images relationship detection and Visual Genome with our large models. 
    \item \textbf{Image feature extraction for downstream vision-language tasks.} We provide a easy-to-use pipeline to extract object and relation features from our pretrained models. The features exacted from our large models, when used as the input of downstream vision-language models, achieve SOTA performance on seven major vision-language tasks~\cite{zhang2021vinvl}.
\end{itemize}

Besides presenting major improvements in our code base (Section~\ref{sec:repoimprov}) and benchmarking results on VG and OI datasets (Section~\ref{sec:benchmarks}), we report a comprehensive ablation study of current models' capability in Section~\ref{sec:abexp}. We study the frequency prior in relationship detection and found that the SOTA models are only slightly better than the frequency prior baseline. We also ablate the error of SGG models into different sources, including object detection error, relation proposal error and relation classification error. 
We hope the study can provide insights to build better SGG models.

% \section{Related Work}
% \label{sec:relatedwork}
% The first widely used codebase is \cite{Zellers_2018}. This codebse contains original implementation of neural motif. However, because this codebase is old and does use some standard computer vision model component \pz{Be more specific here}, it causes some inconvenient for research community to do ablation study and model comparison. 

% Another widely Scene Graph Generation codebase is \url{https://github.com/KaihuaTang/Scene-Graph-Benchmark.pytorch}. This codebase~\cite{tang2020sggcode} is tightly coupled with the object detection model from \cite{massa2018mrcnn}, i.e., relationship prediction module is an additional module of roi-heads in object detection. Different from this tightly-coupled choice, we choose to decouple the object detection model and the relationship detection model.

\section{Major improvements in SGG methods.}
\label{sec:repoimprov}

In our code base, we have re-implemented five widely used SGG algorithms, i.e, IMP \cite{Xu_2017}, MSDN \cite{Li_2017}, GRCNN \cite{Yang_2018}, NM \cite{Zellers_2018} and RelDN \cite{zhang2019graphical}, with two major improvements as described below.

\begin{figure}[t!]
\centering
\includegraphics[width=0.98\columnwidth]{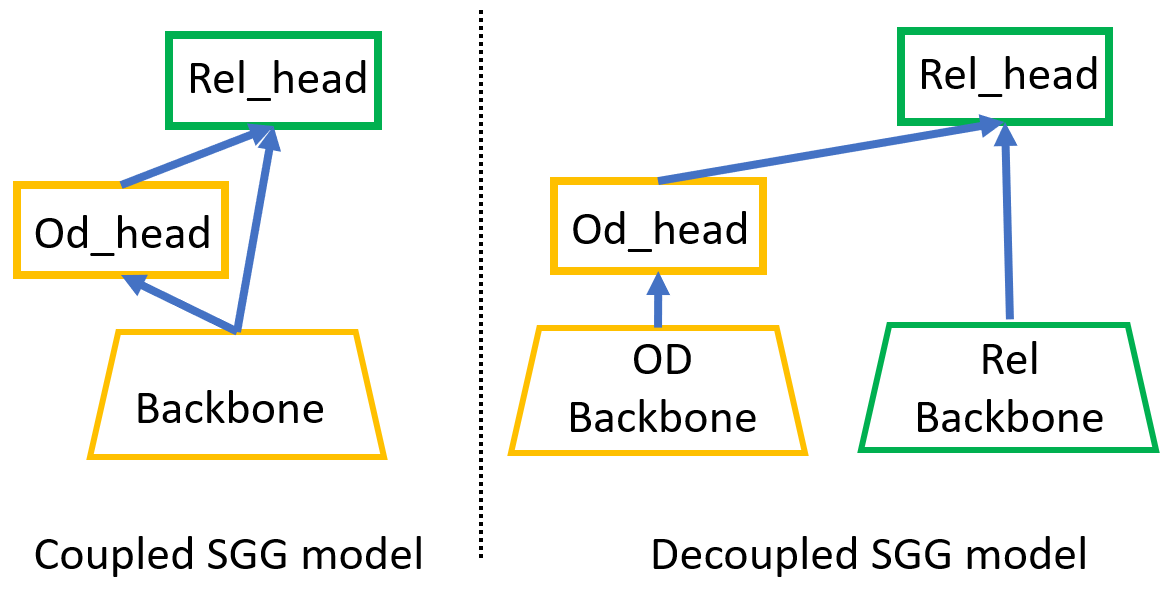}
% \vspace{-3mm}
\caption{Illustration of coupled and decoupled SGG model architecture.}
\label{fig:model_arc}
\vspace{-4mm}
\end{figure}

\subsection{Decouple object and relationship detection}
\label{subsec:decouplearch}
Most existing SGG models use a two-stage model architecture (object detection first and then relationship prediction), e.g., \cite{Xu_2017,Li_2017,Yang_2018,Zellers_2018}, where the relationship detection module extracts relation features from the same backbone of the object detection module, as shown in Figure~\ref{fig:model_arc} Left. The object detector (the yellow part) is pre-trained with the object detection task. To train the relationship detection module (the green part), one either (1) trains only the parameters in the relationship detection module, resulting in sub-optimal relationship detection performance, or (2) trains the full network with both object and relationship detection losses in a multi-task learning fashion, typically resulting in OD performance drop compared to the pre-trained object detector. This tightly coupled model architecture also makes SGG models difficult to make use of off-the-shelf object detectors, such as calling a Cloud service for object detection.

To resolve these issues, our code base supports {\bf both the coupled and decoupled} SGG model architecture. As shown in Figure~\ref{fig:model_arc} Right, the decoupled model introduces a separate relation backbone for relationship detection, which enables users to use their customized detection models/results.  
In our current implementation, we use \cite{massa2018mrcnn} as our object detector, but the decoupled model can receive detection results from any object detectors. In Table~\ref{tab:modelarch}, we show that this decoupled model architecture outperforms the coupled counterpart. The extra computational and memory cost from the relation backbone is minor, because the major cost of an SGG model is from its relation head.  
% In the future, we can even replace our object detector with other framework, like \cite{wu2019detectron2}. 
% Besides, we also provide functionality of extracting object level/relation level features, which can be used in downstream tasks.

\subsection{No refinement of object detection results}
\label{subsec:refine}
It is a common practice in training SGG models that the relationship detection module also learns to output refined object detection results, which either only refines the object classification (labels and confidences), e.g., \cite{Xu_2017,Li_2017,Yang_2018}, or refines both object classification and box regression, e.g., \cite{Zellers_2018}. This refinement introduces extra OD parameters and an OD loss in the relationship detection module. The hope is that by jointly training OD and relationship detection, the OD performance can be improved. However, we found that it is not the case. The refined OD results are typically worse than the pre-trained OD results. As shown in Table~\ref{tab:onlinelabelexps}, the worse and constantly-changing OD results also make it more difficult to train the relationship module, resulting in worse relationship detection performance. Therefore, we do not refine the object detection results in the relationship module. 

% \subsection{Large backbone pre-trained on large merged OD }

\section{SGG Benchmarks on VG and OI}
\label{sec:benchmarks}

\subsection{Datasets and evaluation metrics}
Our code base provides a generic SGG dataset class that supports both VG and OI datasets. It is easily adaptable to customized SGG datasets, too. We listed the main features of these two datasets and their evaluation metrics as below.

\textbf{Visual Genome Scene Graph Generation}. The Visual Genome dataset for Scene Graph Generation, introduced by IMP~\cite{Xu_2017}, contains 150 object and 50 relation categories. We follow their train/val splits.
The evaluation codes are adopted from IMP~\cite{Xu_2017} and NM~\cite{Zellers_2018}. The evaluation metrics are based on Recalls of three settings:  (1) {\bf predicate classification (predcls)}: given a ground truth set of object boxes and labels, predict edge labels, (2) {\bf scene graph
classification (sgcls)}: given ground truth boxes, predict box labels and edge labels and (3) {\bf scene graph detection (sgdet)}: predict boxes, box labels and edge labels. 
% The difference is that we iterate over ground truth annotations, so that even if predictions miss some images, we can guarantee the final recall is calculated correctly. 
    
\textbf{Open Images Visual Relationship Detection}. We follow Open Images Challenge 2018 Visual Relationship Detection Track. We use official recommended validation subset to split the data. The train and val sets contains 53,953 and 3,234 images respectively. There are 9 different relationships, 57 different object classes and 5 different attributes. We adopt the evaluation code from Open Images Visual Relationship Detection Challenge 2018 official evaluation. The metrics are based on detection average precision/recall. Following RelDN, we scale each predicate category by their relative ratios in the val set, which we refer to as the weighted mAP (wmAP), to solve extreme predicate class imbalance problem. The final evaluation score is calculated as \(Score = 0.2 * recall@50 + 0.4 * wmAP(Triplet) + 0.4 * wmAP(Phrase) \).

% \subsection{Evaluation Metric}
% Our codebase provides two set of different evaluation metrics:
% \begin{itemize}
%     \item \textbf{Visual Genome}. The evaluation codes are adopted from Neural Motif and Iterative message passing. The metrics are based on Recalls of three settings: predcls, sgcls and sgdet. 
%     The difference is that we iterate over ground truth images, so that even if predictions miss some images, we can guarantee the final recall is calculated correctly. 
%     \item \textbf{Open Images Visual Relationship Detection}. We adopt the evaluation code from Open Images Visual Relationship Detection Challenge 2018 official evaluation. The metrics are based on detection average precision/recall.
%     Follow RelDN, we scale each predicate category by their relative ratios in the val set, which we refer to as the weighted mAP (wmAP), to solve extreme predicate class imbalance problem.
% \end{itemize}

\subsection{SGG methods and frequency baselines}
We benchmark five scene graph generation methods, including iterative message passing (IMP)~\cite{Xu_2017}, multi-level scene detection networks (MSDN)~\cite{Li_2017}, graph R-CNN (GRCNN)~\cite{Yang_2018}, Neural Motif~\cite{Zellers_2018} and RelDN~\cite{zhang2019graphical}. 

Since object labels are highly predictive of relationship labels, we additionally include two simple yet strong frequency baselines (first proposed in \cite{Zellers_2018}) built from training set statistics. The first, ``Freq", gives the most likely relation type given the object and subject labels, based on training set statistics. The second, ``Freq+Overlap", additionally requires that the two boxes have overlap in order for the pair to count as a valid relation.

\subsection{Benchmarking Results}

We present benchmarking results on Open Images Visual Relationship Detection in Table \ref{table:1}. 
% We also benchmark results in Predicate Detection mode in table \ref{table:8}. 
The object detection model is a ResNeXt152-FPN detector trained on OpenImagesVRD dataset. Its performance is 76.11 mAP@0.5 over 57 object classes used in relation detection. The RelDN performs the best. The ``Freq+Overlap" baseline performs very competitively, better or as good as all methods except RelDN. Thanks to the large and strong ResNeXt152 backbone, our new RelDN result sets a new SOTA performance on this OI VRD task.

\begin{table*}
% \begin{scriptsize}
\begin{center}
\resizebox{\linewidth}{!}{
\begin{tabular}{ c|c|c|c|c|c|c|c|c }
\toprule
Model & Score & Recall@50 & wmAP(Triplet) & mAP(Triplet) & wmAP(Phrase) & mAP(Phrase) & Triplet proposal recall & Phrase proposal recall \\ [0.11ex]
\midrule
% IMP, no bias & 39.71 & 71.64 & 30.56 & 36.47 & 32.90 & 40.61 & 72.57 & 75.87 \\ 
% \hline
% MSDN, bias & 39.42 & 71.48 & 30.22 & 34.49 & 32.58 & 38.71 & 72.45 & 75.62 \\ 
Neural Motif & 39.48 & 72.54 & 29.35 & 29.26 & 33.10 & 35.02 & 73.64 & 78.70 \\ 
\hline
MSDN & 39.64 & 71.76 & 30.40 & 36.76 & 32.81 & 40.89 & 72.54 & 75.85 \\ 
\hline
IMP & 40.01 & 71.81 & 30.88 & 45.97 & 33.25 & 50.42 & 72.81 & 76.04 \\ 
\hline
Freq+Overlap & 40.12 & 72.63 & 30.18 & 31.82 & 33.81 & 37.08 & 74.81 & 78.26 \\ 
\hline
GRCNN & 43.54 & 74.17 & 34.73 & 39.56 & 37.04 & 43.63 & 74.11 & 77.32 \\ 
\hline
RelDN & 51.08 & 75.40 & 40.85 & 44.24 & 49.16 & 50.60 & 78.74 & 90.39 \\ 
\bottomrule
\end{tabular}
}
\end{center}
% \end{scriptsize}
\caption{Open Images Scene Graph Detection results on val set. All models share the same object detector, which is an ResNeXt152-FPN detector trained on 57 object classes used in this dataset. All models are inferenced in unconstrained mode (each subject and object pair can have at most 2 predicates).}
\label{table:1}
\end{table*}

\begin{comment}
\begin{table*}
% \begin{scriptsize}
\begin{center}
\resizebox{\linewidth}{!}{
\begin{tabular}{ c|c|c|c|c|c|c|c|c|c }
\toprule
Model & Recall@50 & wmAP(Triplet) & mAP(Triplet) & wmAP(Phrase) & mAP(Phrase) & \shortstack{Triplet \\ proposal \\ recall} & \shortstack{Phrase \\ proposal \\ recall} & \shortstack{Predicate \\ classification \\ accuracy} \\ [0.11ex]
\midrule
IMP, no bias & 99.16 & 95.67 & 89.88 & 90.89 & 85.63 & 95.74 & 92.45 & 74.82 \\ 
\hline
IMP, bias & 99.63 & 96.47 & 92.67 & 91.34 & 87.76 & 95.74 & 92.45  & 76.21 \\ 
\hline
MSDN, no bias & 99.38 & 96.17 & 84.23 & 91.03 & 79.52 & 95.74 & 92.45 & 74.94 \\ 
\hline
MSDN, bias & 99.72 & 96.71 & 93.05 & 91.58 & 88.14 & 95.74 & 92.45 & 78.55 \\ 
\hline
Neural Motif, bias & 99.78 & 97.36 & 93.95 & 91.92 & 88.97 & 95.74 & 92.45 & 86.69\\ 
\hline
GRCNN, bias & 99.80 & 96.32 & 90.04 & 90.82 & 85.18 & 95.74 & 92.45 & 77.34 \\ 
\hline
RelDN & 99.53 & 94.53 & 79.09 & 89.74 & 74.83 & 95.74 & 92.45 & 72.33 \\ 
\hline
Freq+Overlap & 99.89 & 97.19 & 82.44 & 91.80 & 77.21 & 95.74 & 92.45 & 95.31 \\ 
\bottomrule
\end{tabular}
}
\end{center}
% \end{scriptsize}
\caption{Open Images Predicate Detection results on val set. All models' detection are get from a Faster-RCNN with ResNeXt152FPN backbone finetuned on 57 object classes used in this dataset. All models are inferenced in unconstrained mode.}
\label{table:8}
\end{table*}
\end{comment}

We present benchmarking results on Visual Genome in Table \ref{table:2}. The object detection model is a ResNet50-FPN detector trained on Visucal Genome. Its performance is 28.15 mAP@0.5 over 150 object classes. RelDN is still the best performer. All the other methods achieve similar performance as the ``Freq+Overlap" baseline.

\begin{table*}
% \begin{scriptsize}
\begin{center}
\resizebox{\linewidth}{!}{
\begin{tabular}{ c|c|c|c|c|c|c|c|c|c }
\toprule
Model & sgdet@20 & sgdet@50 & sgdet@100 & sgcls@20 & sgcls@50 & sgcls@100 & predcls@20 & predcls@50 & predcls@100 \\ [0.11ex]
\midrule
% IMP, no bias & 19.8 & 27.5 & 33.0 & 28.0 & 33.4 & 35.1 & 44.9 & 54.8 & 57.8 \\ 
% \hline
% MSDN, no bias & 21.0 & 28.3 & 33.5 & 28.2 & 33.4 & 35.0 & 46.0 & 55.0 & 57.7 \\ 
% \hline
% Neural Motif, no bias & 21.0 & 28.6 & 33.8 & 29.2 & 34.1 & 35.5 & 51.0 & 60.2 & 62.3 \\ 
% \hline
Neural Motif & 21.8 & 30.1 & 33.8 & 30.2 & 35.1 & 36.5 & 52.1 & 61.2 & 63.2 \\ 
\hline
Freq+Overlap & 21.6 & 29.0 & 34.0 & 31.9 & 35.7 & 36.6 & 53.4 & 60.5 & 62.1 \\ 
\hline
IMP & 21.7 & 29.3 & 34.5 & 29.2 & 33.9 & 35.3 & 48.8 & 57.6 & 59.9 \\ 
\hline
% GRCNN, bias & 22.9 & 30.1 & 34.8 & 30.5 & 34.9 & 36.2 & 52.1 & 59.9 & 61.8 \\ 
% \hline
MSDN & 22.4 & 30.0 & 35.3 & 29.7 & 34.4 & 35.9 & 51.2 & 59.6 & 61.6 \\ 
\hline
GRCNN & 22.9 & 30.1 & 34.8 & 30.5 & 34.9 & 36.2 & 52.1 & 59.9 & 61.8 \\ 
\hline
RelDN & 24.0 & 32.4 & 37.8 & 31.9 & 35.7 & 36.6 & 54.0 & 60.9 & 62.5 \\ 
\bottomrule
\end{tabular}
}
\end{center}
% \end{scriptsize}
\caption{Visual Genome Scene Graph Detection results on val set. All models share the same object detector, which is a ResNet50-FPN detector. All models are evaluated in constrained mode (each subject and object pair can only have 1 predicate).}
\label{table:2}
\end{table*}

\section{Ablation Experiments}
\label{sec:abexp}
In order to justify our two major changes to existing SGG methods and understand the limitation of these methods, we perform the following ablation study on the OI VRD task.

\noindent\textbf{Coupled vs decoupled SGG model architecture.}
As mentioned in Section~\ref{subsec:decouplearch}, our code base supports both the coupled and decoupled model architecture. In Table~\ref{tab:modelarch}, we show that the decoupled arch outperforms the coupled arch. Table~\ref{tab:modelarch} also shows that the coupled arch with full model trained by both object and relation detection losses performs the worst. This indicates that the relationship detection task is not helping object detection, but instead making its performance worse.  
% We tried 3 design options: 1. train object detector and relation detector end to end, led to degrade in object detection ; 2. freeze object detector and using its backbone to extract features in relation detector; 3. freeze object detector and train relation detector with a separated backbone. Among these 3 design choices, the 3rd option performs consistently better on all SGG algorithms. See Table \ref{tab:modelarch} for experiment details. Therefore, we decided to use a separate backbone in relation detector, which also matches our design choice of decoupling object detection and relationship detection.

\begin{table*}[h!]
% \begin{tiny}
\begin{center}
\resizebox{\linewidth}{!}{
\begin{tabular}{ c|c|c|c|c|c|c|c|c|c }
\toprule
OD & Relation & score & Recall@50 & wmAP(Triplet) & mAP(Triplet) & wmAP(Phrase) & mAP(Phrase) & \shortstack{Triplet \\ proposal \\ recall} & \shortstack{Phrase \\ proposal \\ recall} \\ [0.11ex]
\midrule
predicted objects & ``Freq+Overlap" & 40.12 & 72.63 & 30.18 & 29.35 & 33.81 & 34.05 & 77.10 & 89.92 \\ 
predicted objects & model relation & 46.96 & 74.42 & 38.78 & 41.38 & 41.41 & 44.68 & 77.51 & 89.98 \\ 
\midrule
gt objects & ``Freq+Overlap" & 78.94 & 93.38 & 77.87 & 73.89 & 72.81 & 68.67 & 93.87 & 94.73 \\ 
gt objects & model relation & 86.41 & 93.29 & 86.46 & 85.36 & 82.93 & 81.76 & 93.84 & 97.32 \\ 
\midrule
gt objects & gt pairs +  ``Freq+Overlap" & 95.58 & 99.89 & 97.21 & 84.72 & 91.80 & 79.66 & 100 & 100 \\ 
gt objects & gt pairs + predicted relation & 94.21 & 99.86 & 95.22 & 84.67 & 90.38 & 80.43 & 100 & 100  \\ 
\bottomrule
% groundtruth boxes, groundtruth pairs, Freq+Overlap relation, constrained & 87.79 & 83.36 & 75.25 & 83.36 & 75.25 & 100 & 100 & 87.79 \\ 
% \hline
% groundtruth boxes, groundtruth pairs, model relation, constrained & 88.88 & 83.00 & 75.57 & 83.06 & 75.82 & 100 & 100 & 88.88 \\ 
% \hline
\end{tabular}
}
\end{center}
% \end{tiny}
\caption{We use the pre-trained RelDN model from \cite{zhang2019graphical} for this ablation study. The mAP of object detector (ResNeXt101-FPN) is 71.33. When using predicted bounding box, the bbox pair recall (regardless of predicate) are 86.94(Triplet) and 95.17(Phrase).}
\label{table:3}
\end{table*}

\begin{table}[ht]
\begin{center}
%\scriptsize
\begin{tabular}{ c|c }
\toprule
 Model & Recall@50 \\
\midrule
Coupled, only Rel\_head is trained &  74.31  \\ 
Coupled, full model is trained &  66.47  \\ 
%  Train detector and relation head at the same time, with seperated backbone & 73.71  \\ 
Decoupled & 75.40  \\ 
\bottomrule
\end{tabular}
\end{center}
\vspace{-4mm}
\caption{Comparison of coupled and decoupled models. All these experiments are conducted on OI VRD Dataset with ResNeXt152-FPN as backbone and RelDN as relation head.}
\label{tab:modelarch}
\vspace{-4mm}
\end{table}

\begin{table}[ht]
\begin{center}
%\scriptsize
\begin{tabular}{ c|c|c }
\toprule
 Model & refined & not refined \\
\midrule
IMP & 69.90 & 71.43 \\ 
% IMP, w/o online label & 71.43  \\ 
MSDN & 68.08 & 71.36 \\ 
% MSDN, w/o online label & 71.36  \\ 
Neural Motif & 71.46 & 72.68  \\ 
% Neural Motif, w/o online label & 72.68  \\ 
\bottomrule
\end{tabular}
\end{center}
\vspace{-4mm}
\caption{Comparison of using object label predicted by relation head (refined object detection) and label predicted by pre-trained object detector. All models are trained on OI VRD dataset with ResNeXt152-FPN backbone and the reported numbers are Recall@50.}
\label{tab:onlinelabelexps}
\vspace{-3mm}
\end{table}

\noindent\textbf{Should OD be refined or not?}
In Table~\ref{tab:onlinelabelexps}, we show that it is more desirable to directly use the pre-trained OD results in the relationship detection module, instead of further refining them.

\noindent\textbf{Ablate SGG error into different sources.}
In Table \ref{table:3}, we ablate SGG errors into three sources: errors from object detection (predicted objects vs ground-truth objects), errors from relation proposal (model's proposed pairs vs ground-truth pairs), and errors from relation classification.
We use two models for this experiment: the pre-trained RelDN model (ResNeXt101-FPN) from \cite{zhang2019graphical} and the ``Freq+Overlap" baseline. We can see that the majority of SGG error comes from the object detection, because the score jumps from 40.96 to 86.41 by replacing the predicted objects with the ground-truth objects. The relation proposal is the second major source of SGG error source, because the score jumps from 86.41 to 94.21 by further replacing the proposed pairs with the ground-truth pairs. Finally, the classification error contributes to the gap from 94.21 to 100. Table \ref{table:3} also shows that the ``Freq+Overlap" baseline is even better than the RelDN model in terms of relation classification, but worse in terms of relation proposals. 

\noindent\textbf{The barrier of frequency/language prior.}
As discussed in the last paragraph, the models' relation classification capability is almost the same as the "Freq+Overlap" baseline on OI VRD. Since VG task has more predicate classes (50) compared with OI VRD (10), it would be more difficult for the "Freq+Overlap" baseline. Therefore, we use models trained on VG task to further study if SGG methods only learned the frequency prior from training data. In Table~\ref{table:vg_model_ensemble}, we first compute every method's relation classification accuracy (the diagonal part) and their pair-wise perfect ensemble accuracy (the lower-diagonal part). As the diagonal shows, all the methods has nearly the same and are slightly worse than the "Freq+Overlap" baseline. However, all the methods are to some degree different from the "Freq+Overlap" baseline because the pair-wise ensemble models (the last row) are clearly better than those single models (the diagonal). 
% Table~\ref{table:vg_force_relation} shows that there's only small difference between different models in terms of predicate classification capability. 
To further quantify the similarity of two SGG models (including the "Freq+Overlap" baseline), We calculated their pair-wise similarity based on models' predictions in Table~\ref{table:vg_model_similarity}. From these results, we can see the model predictions similarity range from 70.89\% to 84.17\%. 
% We also did perfect ensembles for each pair of these models' predictions, please check Table~\ref{table:vg_model_ensemble}. 
From these results, we conclude that the difference in terms of relation classification capability between learned SGG models and the "Freq+Overlap" baseline are very small. To further improve SGG performance, new SGG methods should learn more visual signals different from the frequency prior. 

% We also compared different frequency prior. One is calculated directly from training data. The other is calculated from training data that remove non-overlapping bounding box pairs. Please check at table \ref{table:4}.

% \begin{table}
% \begin{center}
% \begin{tabular}{ c|c|c|c }
% \toprule
% Model & Recall@20 & Recall@50 & Recall@100 \\
%  \midrule
% IMP & 64.31 & 64.66 & 64.66 \\ 
% MSDN & 65.28 & 65.62 & 65.62 \\ 
% Neural Motif & 65.43 & 65.77 & 65.77 \\ 
% GRCNN & 65.34 & 65.70 & 65.70 \\ 
% RelDN & 66.18 & 66.55 & 66.55 \\ 
% Freq+Overlap & 66.14 & 66.47 & 66.47 \\ 
% \bottomrule
% \end{tabular}
% \end{center}
% \vspace{-4mm}
% \caption{Evaluation results on Visual Genome under force relation mode (using ground truth bbox and bbox pairs, different from Predcls, which only uses ground truth bbox). For IMP, GRCNN, MSDN, we only evaluate model trained using bias.}
% \label{table:vg_force_relation}
% \vspace{-3mm}
% \end{table}

\begin{table}[tp!]
\begin{center}
\resizebox{\linewidth}{!}{
\begin{tabular}{ c|c|c|c|c|c|c }
\toprule
Model & IMP & MSDN & Neural Motif & GRCNN & RelDN & Freq+Overlap \\ [0.11ex]
 \midrule
IMP & 63.85 &  &  &  &  &  \\ 
MSDN & 69.85 & 64.82 &  &  &  &  \\ 
Neural Motif & 67.71 & 70.27 & 65.21 &  &  &  \\ 
GRCNN & 70.61 & 68.51 & 70.57 & 64.80 &  &  \\ 
RelDN & 71.14 & 70.03 & 70.67 & 69.99 & 66.12 &  \\ 
Freq+Overlap & 70.38 & 69.36 & 69.69 & 69.76 & 69.03 & 66.23 \\ 
\bottomrule
\end{tabular}
}
\end{center}
\vspace{-4mm}
\caption{Visual Genome models predicate predictions pair-wise ensemble accuracy ($\frac{TP+TN}{TP + TN + FP + FN}$). We assume a perfect ensemble setting where the ensemble model is correct as long as there is a correct prediction in the ensemble.}
\label{table:vg_model_ensemble}
\vspace{-3mm}
\end{table}

\begin{table}[!htbp]
\begin{center}
\resizebox{\linewidth}{!}{
\begin{tabular}{ c|c|c|c|c|c|c }
\toprule
Model & IMP & MSDN & Neural Motif & GRCNN & RelDN & Freq+Overlap \\ [0.11ex]
 \midrule
IMP & 100 & & & & \\ 
MSDN & 73.83 & 100 & & & \\ 
Neural Motif & 84.17 & 74.53 & 100 & & & \\ 
GRCNN & 71.06 & 80.12 & 73.59 & 100 & &  \\ 
RelDN & 70.89 & 75.98 & 75.15 & 76.86 & 100 & \\ 
Freq+Overlap & 74.49 & 79.17 & 80.30 & 78.21 & 83.56 & 100 \\ 
\bottomrule
\end{tabular}
}
\end{center}
\vspace{-4mm}
\caption{Visual Genome model predictions similarity comparison under force relation mode. The similarity is defined as $\frac{\text{\# matched triplets}}{\text{\# total unique triplets}}$, triplet is defined as $(subj\_instance, predicate, obj\_instance)$.}
\label{table:vg_model_similarity}
\vspace{-3mm}
\end{table}

\section{Feature Extraction Pipelines}
\label{sec:feature_extraction}
Our SGG benchmark supports feature extraction for downstream computer vision and vision-language tasks. Given an image, we can extract both object-level features and relation-level features. 

Object-level features are extracted from the object detector. Each bounding box has a feature vector. Relation-level features are extracted from the scene graph detector. Each relation pair has a feature vector. 

\clearpage
{\small
\bibliographystyle{ieee}
\bibliography{paperbib}
}

\end{document}